\documentclass[sigconf,10pt,nonacm]{acmart}
\AtBeginDocument{%
  }

\setcopyright{cc}
\setcctype{by-nc-sa}
\copyrightyear{2026}
\acmYear{2026}

\usepackage{graphicx}      
\usepackage{amsmath}       
\usepackage{booktabs}      
\usepackage{algorithm}     
\usepackage{algpseudocode} 
\usepackage{kotex}         
\usepackage{subfig}        
\usepackage{subcaption}    
\usepackage{makecell}      
\usepackage{multirow}      
\usepackage[framemethod=TikZ]{mdframed} 
\usepackage{diagbox}       
\usepackage{pifont}        
\usepackage{tcolorbox}     
\usepackage{xcolor}        
\usepackage{enumitem}      
\usepackage{tabularx}      
\usepackage{array}         
\usepackage{threeparttable}
\newcolumntype{C}[1]{>{\centering\arraybackslash}p{#1}}
\newcommand{\pmark}{\ding{108}} 
\newcommand{\prtmark}{\ding{115}} 

\begin{document}

\title{Why Do AI Agents Systematically Fail at Cloud Root Cause Analysis?}
\settopmatter{authorsperrow=4}

\author{Taeyoon Kim}
\email{tykim7@hanyang.ac.kr}
\orcid{0009-0003-0705-0713}
\affiliation{%
  \department{Dept. of Data Science}
  \institution{Hanyang University}
  \city{Seoul}
  \country{Republic of Korea}
}
\authornote{Both authors contributed equally to this work.}

\author{Woohyeok Park}
\email{woohyeok@hanyang.ac.kr}
\orcid{0009-0008-9401-0416}
\affiliation{%
  \department{Dept. of Data Science}
  \institution{Hanyang University}
  \city{Seoul}
  \country{Republic of Korea}
}
\authornotemark[1]

\author{Hoyeong Yun}
\email{hy.yun@okestro.com}
\orcid{0009-0000-1343-7023}
\affiliation{%
  \department{Intelligent Cloud Lab}
  \institution{OKESTRO Co., Ltd.}
  \city{Seoul}
  \country{Republic of Korea}
}

\author{Kyungyong Lee}
\email{kyungyong@hanyang.ac.kr}
\orcid{0000-0003-0312-4386}
\affiliation{%
  \department{Dept. of Data Science}
  \institution{Hanyang University}
  \city{Seoul}
  \country{Republic of Korea}
}
\authornote{Corresponding author}

\begin{abstract}
Failures in large-scale cloud systems incur substantial financial losses, making automated Root Cause Analysis (RCA) essential for operational stability. Recent efforts leverage Large Language Model (LLM) agents to automate this task, yet existing systems exhibit low detection accuracy even with capable models, and current evaluation frameworks assess only final answer correctness without revealing why the agent's reasoning failed. This paper presents a process-level failure analysis of LLM-based RCA agents. We execute the full OpenRCA benchmark across five LLM models, producing 1,675 agent runs, and classify observed failures into 12 pitfall types across \textit{intra-agent} reasoning, \textit{inter-agent} communication, and \textit{agent-environment} interaction. Our analysis reveals that the most prevalent pitfalls, notably hallucinated data interpretation and incomplete exploration, persist across all models regardless of capability tier, indicating that these failures originate from the shared agent framework rather than from individual model limitations. Controlled mitigation experiments further show that prompt engineering alone cannot resolve the dominant pitfalls, whereas enriching the \textit{inter-agent} communication protocol reduces communication-related failures by up to 15 percentage points. The pitfall taxonomy and diagnostic methodology developed in this work provide a foundation for designing more reliable autonomous agents for cloud RCA.
\end{abstract}

\keywords{Root Cause Analysis, AI Agent, Cloud Operations, AIOps}

\maketitle

\section{Introduction}
Failures in large-scale distributed web services result in significant downtime and financial losses, rendering automated Root Cause Analysis (RCA) essential for service reliability. RCA remains challenging due to complex microservice dependencies and the need to correlate heterogeneous telemetry data across metrics, logs, and traces. The advent of LLMs has accelerated multi-agent systems for automating this task, with frameworks like OpenRCA~\cite{openrca} providing representative benchmarks and baseline agent frameworks.

However, the practical performance of these baseline agents remains very low (Table~\ref{tab:openrca-baseline}), with overall perfect accuracy ranging from 3.9\% to 12.5\% across five models spanning different capability tiers. Furthermore existing evaluation frameworks~\cite{aiopslab, agent-as-a-judge, rca-eval} assess only final answer correctness without revealing why the agent's reasoning or collaboration fails. Without such process-level understanding, practitioners cannot determine whether to invest in stronger models, better prompts, or redesigned agent frameworks.

This paper addresses this gap through a systematic, process-level failure analysis. We execute the full OpenRCA benchmark across five LLM models, producing 1,675 agent runs, and classify the observed failures into 12 pitfall types across \textit{intra-agent} reasoning, \textit{inter-agent} communication, and \textit{agent-environment} interaction. We further conduct mitigation experiments for each category, finding that structural modifications to the agent framework yield measurable improvements while prompt-level interventions alone do not.

The main contributions of this work are as follows.
\begin{itemize}[topsep=0pt,leftmargin=2em]
\item A process-level analysis methodology and a taxonomy of 12 pitfall types that enables systematic diagnosis of RCA agent failures beyond outcome-based evaluation.
\item An empirical analysis showing that the dominant failure modes are shared across all models and originate from the agent framework rather than model limitations.
\item Controlled mitigation experiments showing that enriched \textit{inter-agent} communication produces consistent performance gains, while prompt-level interventions alone do not resolve the identified pitfalls.
\end{itemize}

\section{LLM Agent for RCA}
\label{sec:agent-for-rca}
\begin{table}[t]
  \centering
  \caption{Accuracy comparison of the baseline OpenRCA agent across three service domains}
  \label{tab:openrca-baseline}
  \small
  \begin{threeparttable}
    \resizebox{\linewidth}{!}{
      \begin{tabular}{l|cc|cccccc}
        \toprule
        & \multicolumn{2}{c|}{\textbf{Overall\,(\%)}} &
        \multicolumn{2}{c}{\textbf{Telecom\,(\%)}} &
        \multicolumn{2}{c}{\textbf{Bank\,(\%)}} &
        \multicolumn{2}{c}{\textbf{Market\,(\%)}} \\
        \cmidrule(lr){2-3}\cmidrule(lr){4-5}\cmidrule(lr){6-7}\cmidrule(lr){8-9}
        \textbf{Model} & \pmark & \prtmark & \pmark & \prtmark & \pmark & \prtmark & \pmark & \prtmark \\
        \midrule
        Gemini 2.5 Pro        & \textbf{12.5} & \textbf{22.4} & 11.8 & 13.7 & \textbf{19.9} & \textbf{19.9} & \textbf{6.1} & \textbf{27.7} \\
        GPT-5 mini            &  8.4          & 21.5          & \textbf{15.7} & \textbf{21.6} & 11.8 & 16.2 &  2.7 & 26.4 \\
        GPT-OSS 120B          &  6.9          & 12.2          &  9.8 & 15.7 &  7.4 & 11.8 &  5.4 & 11.5 \\
        Solar Pro 2           &  5.7          & 15.8          &  9.8 & 11.8 &  6.6 & 15.4 &  3.4 & 17.6 \\
        Claude Sonnet 4       &  3.9          & 14.3          &  5.9 &  7.8 &  4.4 & 15.4 &  2.7 & 15.5 \\
        \midrule
        Claude Sonnet 3.5\tnote{*} & 11.3 & 17.3 & -- & -- & -- & -- & -- & -- \\
        \bottomrule
      \end{tabular}
    }
    \begin{tablenotes}[flushleft]
      \footnotesize
    \item[  *]Results from the original RCA-Agent work~\cite{openrca}. \quad\pmark\, Perfect / \prtmark\, Partial
    \end{tablenotes}
  \end{threeparttable}
\end{table}
RCA is the systematic process of identifying the originating source of failure in software systems. Given an observed anomaly, the goal is to determine the faulty component (e.g., a specific microservice), the incident time when the fault first occurred, and the reason for the failure (e.g., memory leak). In practice, these elements are rarely observable directly. A fault in one component propagates through service dependencies and manifests as symptoms in other components, so the engineer must trace the causal chain backward from observed symptoms to the originating source.

This tracing process requires correlating heterogeneous telemetry data across multiple modalities. System metrics such as CPU utilization and memory usage provide time-series signals at the infrastructure level. Application logs record timestamped events and error messages in unstructured text. Distributed traces capture the end-to-end path of individual requests across services, revealing latency and dependency relationships. Each modality provides a partial view of the system state, and effective RCA demands joint analysis across all three. RCA therefore traditionally relies on expert engineers with deep domain knowledge to interpret these signals and localize faults.

Recent work has applied LLM reasoning techniques such as ReAct~\cite{react} and Chain of Thought~\cite{cot} to automate RCA, but embedding large volumes of telemetry data into a single agent's context faces scalability limitations. To address this, recent research~\cite{openrca, rca-flow-of-action} has adopted multi-agent architectures that decompose complex responsibilities across specialized agents, reducing individual agent complexity and enabling programmatic management of large-scale operational tasks.

In agent-based RCA, OpenRCA~\cite{openrca} provides a representative benchmark and baseline agent system. The benchmark comprises 335 failure incidents from three web service domains, namely Telecom (51 cases), Bank (136 cases), and Market (148 cases). Each incident is annotated with a ground truth root cause specifying the faulty component and failure reason, and the dataset covers 73 unique components and 28 distinct failure reasons in total. For each incident, OpenRCA provides three types of telemetry data corresponding to the modalities described above, including system metrics, application logs, and distributed traces. The full dataset totals 68.5GB and over 523 million lines across these modalities.

Alongside the dataset, OpenRCA proposes an agent system for RCA that adopts a Controller-Executor architecture. The Controller performs high-level reasoning and produces natural language instructions for the Executor, which translates them into Python code, executes it over telemetry data, and returns results. We refer to this orchestration layer collectively as the agent framework, distinct from the underlying LLM. Despite these contributions, the detection accuracy remains very low, as shown in Table~\ref{tab:openrca-baseline}. A perfect detection requires all three root cause elements namely component, time, and reason to be correct, while Partial indicates that only a subset is identified. Even the best performing model, Gemini 2.5 Pro, achieves only 12.5\% perfect detection across all 335 incidents, yet existing evaluation frameworks confine assessment to final answer correctness and provide no insight into why the agent's reasoning failed~\cite{aiopslab, agent-as-a-judge}.

\section{Agent Failure Diagnosis Methodology}
\begin{figure}[t!]
  \centering
  \includegraphics[width=0.46\textwidth]{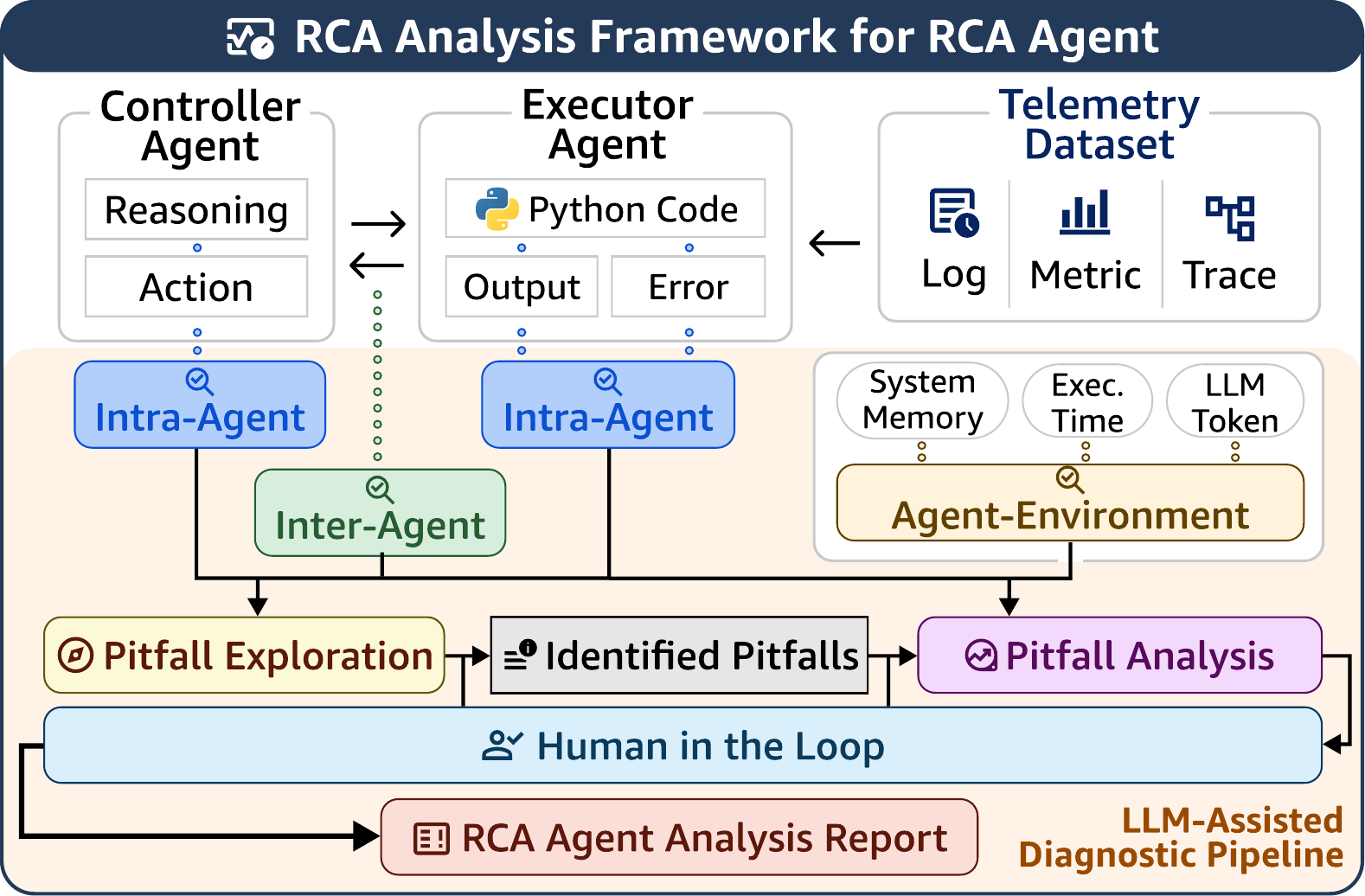}
  \caption{Analysis framework for diagnosing RCA agent failures across three interfaces}
  \Description{Analysis framework for diagnosing RCA agent failures across three interfaces}
  \label{fig:openrca-arch}
\end{figure}

\begin{table*}[t!]
  \centering
  \caption{Pitfall taxonomy with diagnostic criteria}
  \label{tab:pitfall_taxonomy}
  \footnotesize
  \begin{tabularx}{\textwidth}{@{}c|l|l|l@{}}
    \toprule
    \textbf{Category} & \textbf{Pitfall} & \textbf{Core Issue} & \textbf{Diagnostic Question} \\
    \midrule
    \multirow{7}{*}{\textbf{\textit{Intra-Agent}}}
    & Hallucination in Interpretation & Narrative fabrication & Did the agent misinterpret what the data means? \\
    & Incomplete Exploration & Scope narrowing & Did the agent skip a relevant KPI \& Component? \\
    & Symptom-as-Cause & Premature attribution & Did the agent mistake a symptom for the root cause? \\
    & Code generation error & LLM code quality deficit & Did the generated code execute correctly? \\
    & Limited Telemetry Coverage & Source narrowness & Did the agent rely on only one telemetry source? \\
    & Timestamp Error & Temporal misalignment & Did the agent analyze the correct time window? \\
    & No Cross-Validation & Single-hypothesis bias & Did the agent verify findings with alternative sources? \\
    \midrule
    \multirow{3}{*}{\textbf{\textit{Inter-Agent}}}
    & Instruction-Code Mismatch & Implementation gap & Did the code reflect the Controller's intent? \\
    & Meaningless Repetition & Repetitive loop & Did the agent repeat the same failed approach? \\
    & Misattributed Evidence & Opaque handoff & Did the Controller misunderstand what the Executor's results represent? \\
    \midrule
    \multirow{2}{*}{\textit{\textbf{
          \begin{tabular}{@{}c@{}}Agent-\\Environment
    \end{tabular}}}}
    & Out-of-Memory & Resource exhaustion & Did execution terminate due to memory limits? \\
    & Max Step Exhaustion & Budget depletion & Did the agent run out of steps? \\
    \bottomrule
  \end{tabularx}
\end{table*}

\begin{table*}[t!]
\centering
\caption{Distribution of Identified Intra \& Inter Pitfalls (N=1675)}
\label{tab:pitfall_summary}
\footnotesize
\setlength{\tabcolsep}{5pt}
\begin{tabularx}{\textwidth}{c|l|l|@{\hspace{7pt}}c@{\hspace{7pt}}c@{\hspace{7pt}}c@{\hspace{7pt}}c@{\hspace{7pt}}c}
\toprule
\textbf{Category} & \textbf{Pitfall} & \textbf{Overall} & \textbf{Claude Sonnet 4} & \textbf{Solar Pro 2} & \textbf{GPT-5 mini} & \textbf{GPT-OSS 120B} & \textbf{Gemini 2.5 Pro} \\
\midrule
& \textbf{Hallucination in Interpretation} & \textbf{1193 (71.2\%)} & \textbf{267 (79.5\%)} & \textbf{244 (72.8\%)} & \textbf{232 (69.3\%)} & \textbf{226 (67.5\%)} & 224 (66.9\%) \\
& Incomplete Exploration & 1071 (63.9\%) & 247 (73.5\%) & 180 (53.7\%) & 191 (57.0\%) & 224 (66.9\%) & \textbf{229 (68.4\%)} \\
& Symptom-as-Cause & 669 (39.9\%) & 115 (34.2\%) & 117 (34.9\%) & 152 (45.4\%) & 151 (45.1\%) & 134 (40.0\%) \\
& Code generation error & 456 (27.2\%) & 220 (65.5\%) & 46 (13.7\%) & 139 (41.5\%) & 45 (13.4\%) & 6 (1.8\%) \\
& Limited Telemetry Coverage & 451 (26.9\%) & 132 (39.3\%) & 41 (12.2\%) & 105 (31.3\%) & 54 (16.1\%) & 119 (35.5\%) \\
& Timestamp Error & 390 (23.3\%) & 114 (33.9\%) & 98 (29.3\%) & 83 (24.8\%) & 47 (14.0\%) & 48 (14.3\%) \\
\multirow{-7}{*}{\shortstack[c]{\textbf{\textit{Intra-}}\\\textbf{\textit{Agent}}}} & No Cross-Validation & 311 (18.6\%) & 119 (35.4\%) & 45 (13.4\%) & 43 (12.8\%) & 64 (19.1\%) & 40 (11.9\%) \\
\midrule
& Instruction-Code Mismatch & 405 (18.8\%) & 148 (24.8\%) & 51 (22.5\%) & 168 (25.5\%) & 36 (11.8\%) & 2 (0.5\%) \\
& Meaningless Repetition & 179 (8.3\%) & 45 (7.5\%) & 0 (0.0\%) & 116 (17.6\%) & 18 (5.9\%) & 0 (0.0\%) \\
\multirow{-3}{*}{\shortstack[c]{\textbf{\textit{Inter-}}\\\textbf{\textit{Agent}}}} & Misattributed Evidence & 73 (3.4\%) & 38 (6.4\%) & 4 (1.8\%) & 23 (3.5\%) & 8 (2.6\%) & 0 (0.0\%) \\
\bottomrule
\end{tabularx}
\end{table*}

To systematically diagnose the causes of the OpenRCA agent's low detection performance, we developed an analysis framework that classifies failures into three categories according to their architectural origin, providing multiple diagnostic perspectives on whether a failure arises from an individual agent's reasoning, from miscommunication between agents, or from the execution environment. As illustrated in Figure~\ref{fig:openrca-arch}, the three categories correspond to \textit{intra-agent} reasoning within each agent, \textit{inter-agent} communication between the Controller and Executor, and \textit{agent-environment} interaction with the telemetry data and execution runtime.

The analysis covers 335 tasks executed across five models, yielding 1,675 individual agent runs. Each run consists of multiple steps, averaging 11.1 steps per run. Collectively, these runs consumed 609.9 hours of wall clock time, approximately 1.38 billion tokens, and an estimated \$1,394 in cost.

Manually inspecting all runs at this scale is infeasible, so we constructed an analysis pipeline using Claude Code~\cite{anthropic-claude-code} with Opus 4.5 as the analysis agent~\cite{llm-as-judge}. Following the human-in-the-loop principle~\cite{hitl-agent}, we verified every classification before inclusion in the final taxonomy.

We restricted the analysis agent to binary diagnostic questions to minimize bias and improve reproducibility. The pipeline operates in two stages. In the first stage, the agent performs free-form exploration with only a JSON output schema, surfacing diverse failure patterns without predefined categories. In the second stage, we consolidated the collected patterns into mutually exclusive categories and repeated the analysis with binary diagnostic questions such as ``Did the agent skip a relevant KPI \& Component?''. The resulting taxonomy of 12 pitfall types is presented in Table~\ref{tab:pitfall_taxonomy}.

\section{Architectural Pitfalls in RCA Agents}
We selected five reasoning models to represent the current LLM ecosystem, given their increasing adoption for complex agentic tasks. We also considered both capability and inference cost. Claude Sonnet 4, GPT-5 mini, and Gemini 2.5 Pro cover major providers. Solar Pro 2 represents a compact scale, and GPT-OSS 120B serves as an open-weight alternative.

Table~\ref{tab:pitfall_summary} presents the distribution of the identified pitfalls across 1,675 agent executions of five LLM models, sorted by overall frequency. To capture the full spectrum of architectural failure modes, we record every pitfall observed per execution rather than assigning each execution to a single category, as recording only one dominant pitfall would obscure the relative frequency of each failure mode and limit the ability to prioritize architectural improvements. Column percentages therefore sum to more than 100\%.
The following subsections examine each pitfall category in detail.

\subsection{Intra-Agent Pitfalls}
\label{sec:intra-pitfalls}
We group the seven \textit{\textbf{intra-agent}} pitfalls in Table~\ref{tab:pitfall_summary} according to where in the diagnostic process the failure occurs.

\textbf{\textit{Hallucination in Interpretation}}, observed in 71.2\% of executions, spans all stages of this process. Rather than faithfully reading the returned data, the Controller imposes an interpretive narrative onto it, assigning meaning that appears coherent but does not reflect what the values indicate, a pattern consistent with the common generative bias of LLMs. The remaining six pitfalls fall into three categories.

\textit{\textbf{Incomplete Exploration}} and \textit{\textbf{Limited Telemetry Coverage}} reflect a common failure in investigation breadth. \textit{{Incomplete Exploration}} at 63.9\% manifests at component and Key Performance Indicator (KPI) level. Although the OpenRCA framework provides lists of candidate components and KPI families in its prompt, agents routinely ignore entire categories, analyzing CPU metrics exclusively while never querying network KPIs, for example. \textit{{Limited Telemetry Coverage}} at 26.9\% operates at the data modality level. Agents predominantly draw conclusions from metric data alone and rarely examine log or trace sources, even though the framework provides all three for every incident.

\textit{\textbf{Symptom-as-Cause}} and \textit{\textbf{No Cross-Validation}} reflect the opposite failure in the depth of investigation. Even when agents reach a relevant analysis target, they terminate prematurely rather than pursuing the full causal chain. \textit{{Symptom-as-Cause}} at 39.9\% captures cases where the agent treats the first anomaly it encounters as the root cause without tracing further upstream. \textit{{No Cross-Validation}} at 18.6\% reflects a related tendency to accept a single finding without verifying it against alternative telemetry sources.

\textit{\textbf{Code generation error}} and \textit{\textbf{Timestamp Error}} reflect limitations in the agent's execution capability rather than its analytical reasoning. Code generation error at 27.2\% varies sharply, from 1.8\% for Gemini 2.5 Pro to 65.5\% for Claude Sonnet 4, indicating that code generation reliability remains a differentiating factor. \textit{{Timestamp Error}} at 23.3\% arises mainly from timezone misalignment in the OpenRCA dataset, suggesting that such runtime details are better handled through data preprocessing than through prompt instructions.

Each model shows distinct failure profiles. Specifically, Code generation error ranges from 1.8\% for Gemini 2.5 Pro to 65.5\% for Claude Sonnet 4, while \textit{Symptom-as-Cause} rates peak above 45\% for GPT-5 mini and GPT-OSS 120B. Despite these variations, all five models exhibit \textit{{Hallucination in Interpretation}} above 66\% and \textit{{Incomplete Exploration}} above 53\%.

Since the five models differ in provider, capability, and cost tier but share the same agent framework, these uniformly high rates point to the shared framework as the primary bottleneck rather than to any individual model's limitations.

\subsection{Inter-Agent Pitfalls}
\label{sec:inter-pitfalls}
In the OpenRCA system, the Controller and Executor communicate exclusively through natural language summaries, with neither agent having access to the other's internal context. This opaque interface produces three \textit{\textbf{inter-agent}} pitfalls identified through step-level analysis.

\textit{\textbf{Instruction-Code Mismatch}} is the most pervasive, affecting 20 to 26\% of all execution steps for GPT-5 mini, Solar Pro 2, and Claude Sonnet 4. Summarized instructions strip away contextual cues needed to reconstruct the sender's intent~\cite{beyond-context}, causing the Executor to misinterpret the analytical question and generate code that addresses a different one.

\textit{\textbf{Meaningless Repetition}}, observed at 17.6\% for GPT-5 mini, represents an escalated form of the same communication failure. Without shared execution history, the Controller cannot recognize that a previous instruction has already failed and repeats the same directive, entering a loop that consumes the step budget without advancing diagnosis.

\textit{\textbf{Misattributed Evidence}} at 3.4\% originates on the Controller side. Unlike \textit{Hallucination in Interpretation}, where the Controller misreads the data itself, this pitfall stems from the opacity of the Executor's process. The Controller receives only a natural language summary of the Executor's findings without visibility into the code that produced them. When the Executor's implementation deviates from the intended analysis, the Controller has no means to detect the discrepancy and accepts the reported findings at face value, building subsequent reasoning on a flawed evidential basis.

The severity of these pitfalls varies sharply across models. Gemini 2.5 Pro exhibits nearly zero rates across all three categories, while Claude Sonnet 4 and GPT-5 mini record the highest \textit{Instruction-Code Mismatch} at 24.8\% and 25.5\% respectively, with GPT-5 mini also exhibiting the highest \textit{Meaningless Repetition} at 17.6\%. All three pitfalls stem from the information loss inherent to the natural language summary interface, suggesting that enriching the communication channel between agents may reduce these failures.

\subsection{Agent-Environment Pitfalls}
\label{sec:env-pitfalls}
To enhance resource efficiency and reduce latency, the OpenRCA agent system uses a persistent Python kernel that retains variables and data structures in memory across multiple turns, eliminating the need for redundant data loading. However, agents lack awareness of the kernel's accumulated state, and this disconnect produces two environment-level failures.

\textbf{\textit{Out of Memory (OOM)}} failures were detected through the memory profiler integrated during our analysis. In the Bank domain, 2 of 41 scenarios terminated during execution when agents reloaded datasets already in memory or failed to release obsolete variables. These failures are categorical rather than partial, as an OOM crash terminates the entire diagnostic session regardless of reasoning quality.

\textbf{\textit{Max Step Exhaustion}} accounts for 4.1\% of executions overall, with GPT-5 mini at 10.4\% and Claude Sonnet 4 at 8.3\%, while Solar Pro 2 never triggers this pitfall. Whether step exhaustion reflects inefficient execution or more thorough exploration attempts requires a joint analysis of step-level exploration breadth and step consumption.

\section{Mitigating Pitfalls in RCA Agent}
\label{sec:mitigating-pitfalls}
Building on the pitfall analysis, this section examines whether targeted interventions can mitigate each failure category. Rather than proposing a new end-to-end system, we conduct controlled experiments that isolate the effect of a single modification per category, identifying which pitfalls respond to focused changes and which demand fundamental redesign.

\subsection{Intra-Agent: Paradoxes of Prompt Engineering}
\label{sec:paradox-prompt-eng}
The most intuitive response to the \textit{intra-agent} pitfalls is to improve the Controller's prompt. We tested two approaches on Claude Sonnet 4, selected for its lowest baseline accuracy, across 70 Bank domain tasks. Hypothesis-driven prompting~\cite{hypothesis-gen-llm} augments the prompt with a structured template requiring explicit hypotheses for each KPI category, targeting Incomplete Exploration by forcing broader coverage. Pitfall-aware prompting injects descriptions of the frequent pitfalls, instructing the agent to avoid patterns such as skipping network KPIs or accepting the first anomaly as the root cause.

Neither produced a meaningful improvement in root cause identification. Hypothesis-driven prompting did succeed in broadening the exploration scope, with previously ignored KPI categories such as memory utilization appearing as explicit hypotheses. However, the \textit{Hallucination in Interpretation} pitfall persisted at comparable rates. The agent reached the relevant data but continued to impose incorrect interpretations onto it, generating plausible but unfounded narratives regardless of the additional guidance provided. Pitfall-aware prompting exhibited a similar disconnect, with the agent acknowledging pitfall descriptions in its reasoning trace but reproducing the same interpretive failures in practice.

These results demonstrate a clear asymmetry in what prompt engineering can and cannot achieve. While augmented prompts successfully broadened the exploration scope, the Controller continued to fabricate interpretations of the retrieved data comparably, indicating that the hallucination pitfall is not a matter of insufficient guidance but a structural property of the generation process. Mitigating this failure requires architectural intervention, such as external verification modules, rather than further prompt refinement.

\subsection{Inter-Agent: Enriched Communication}
\label{sec:transparent-comm}
To mitigate the information loss inherent to the opaque communication interface, we enriched the protocol so that the Executor returns the generated Python code and complete execution output, including exceptions and stack traces, alongside its natural language summary. Symmetrically, the Executor receives the Controller's full diagnostic analysis, a snippet of the previous execution output, and the overall objective, enabling more closely aligned code generation.

\begin{figure}[t]
  \centering
  \includegraphics[width=0.48\textwidth]{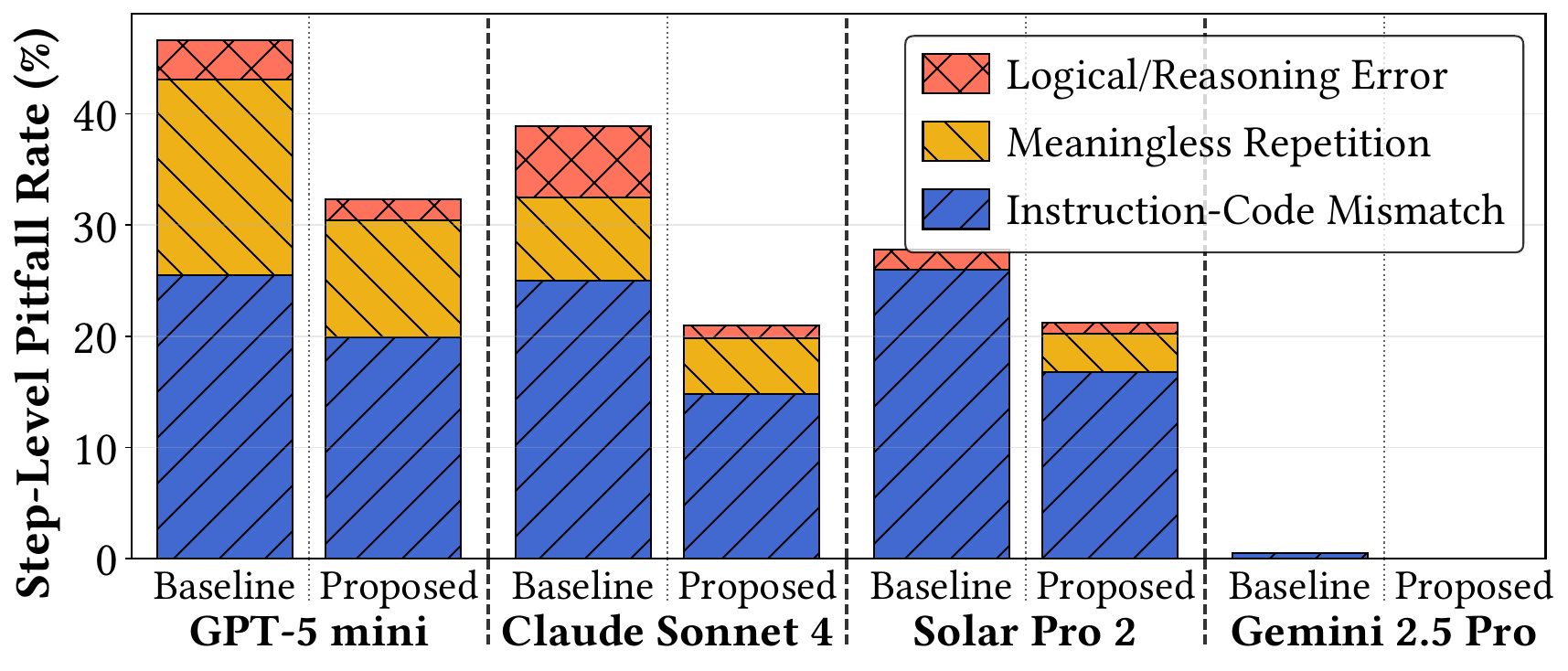}
  \caption{Step-level inter-agent pitfall rates under baseline and enriched communication}
  \label{fig:cmp-blackbox}
\end{figure}

Figure~\ref{fig:cmp-blackbox} compares the pitfall frequency under both protocols. All four models show reduced pitfall rates, with GPT-5 mini and Claude Sonnet 4 exhibiting the largest gains of approximately 14 and 15 percentage points, respectively. Code exposure enables the Controller to directly detect instruction-code discrepancies, while concrete raw error messages prevent the Controller from repeating failed instructions.

These pitfall reductions translate into measurable improvements in RCA performance. In Bank domain experiments, the number of perfect detections increases across all four models under the enriched protocol, with GPT-5 mini rising from 0 to 2 (0.0\% to 4.9\%), Gemini 2.5 Pro from 1 to 3 (2.4\% to 7.3\%), and Solar Pro 2 from 2 to 3 (4.9\% to 7.3\%). 

Although the enriched protocol increases token consumption per step by 24.8\% on average (from 68K to 85K tokens), the reduced number of steps ($-$22.1\%, from 11.9 to 9.2 per run) yields a net decrease of 1.6\% in total tokens and 22.3\% in execution time. By exposing code, errors, and diagnostic context between agents, a simple structural modification to the communication interface can simultaneously improve both accuracy and efficiency, achieving gains that prompt engineering alone could not deliver.

\subsection{Agent-Environment: Robust State Isolation}

As described in Section~\ref{sec:env-pitfalls}, the persistent kernel retains variables across turns, but agents lack awareness of the kernel's accumulated state. When agents reload datasets already in memory or fail to release obsolete variables, the resulting OOM crash terminates the entire diagnostic session regardless of the quality of the reasoning. To prevent this, we integrated a memory watcher that monitors kernel consumption and, upon exceeding a predefined threshold, terminates execution and transmits a structured warning to the Controller, which then generates a more memory-efficient implementation on a restarted kernel. Validation across all models and domains confirmed that this mechanism eliminates all OOM failures observed in the baseline setup.

\section{Conclusion and Future Work}
\label{sec:conclusion}
This paper presented a process-level failure analysis of LLM-based RCA agents that moves beyond outcome-based evaluation to diagnose why agents fail rather than merely how often. Through 1,675 agent runs across five LLM models on the full OpenRCA benchmark, we classified failures into 12 pitfall types spanning \textit{intra-agent} reasoning, \textit{inter-agent} communication, and \textit{agent-environment} interaction. The analysis revealed that \textit{Hallucination in Interpretation} (71.2\%) and \textit{Incomplete Exploration} (63.9\%) persist across all tested models at comparable rates regardless of capability tier, establishing that these dominant failures originate from the shared agent framework rather than from individual model limitations.

Mitigation experiments reinforced this architectural diagnosis. Prompt engineering broadened the scope of investigation but could not suppress the interpretive errors that constitute the most prevalent pitfall. In contrast, enriching the \textit{inter-agent} communication protocol with code and execution output reduced communication-related pitfalls by up to 15 percentage points, improved detection scores across all models, and simultaneously reduced execution time by 22.3\% through more efficient step utilization. A memory watcher that intercepts resource exhaustion before kernel crashes further eliminated all OOM failures observed in the baseline. Together, these results demonstrate that structural modifications to the agent framework are both necessary and effective where prompt-level interventions are not.

This study has several limitations. The mitigation experiments were conducted on the Bank domain subset, the diagnostic pipeline relied on semi-automated classification with human verification that limits scalability, and the generalizability of the pitfall taxonomy to other multi-agent RCA frameworks remains to be validated. Future work will pursue two complementary directions. First, the pitfall taxonomy and binary diagnostic criteria can serve as the basis for a continuous monitoring pipeline that identifies failure patterns across agent runs without manual inspection. Second, the persistent dominance of \textit{Hallucination in Interpretation} despite prompt-level intervention indicates the need for more fundamental architectural changes such as verification modules that cross-check agent interpretations against raw data, structured state sharing, and adaptive task decomposition.

\begin{acks}
  This work was supported by Institute of Information \& communications Technology Planning \& Evaluation (IITP) grant funded by the Korea government (MSIT) (RS-2025-25441560 \& RS-2022-00144309) \& Upstage-AWS AI Initiative Program.
\end{acks}

\bibliographystyle{ACM-Reference-Format}
\bibliography{rca-agent}

\end{document}